\documentclass{article}

\usepackage{arxiv}

\usepackage[utf8]{inputenc}
\usepackage[T1]{fontenc}
\usepackage{hyperref}
\usepackage{url}
\usepackage{booktabs}
\usepackage{amsfonts}
\usepackage{amsmath}
\usepackage{nicefrac}
\usepackage{microtype}
\usepackage{graphicx}
\graphicspath{ {./images/} }

\title{Cross-Prompt Generalization in Detecting AI-Generated Fake News Using Interpretable Linguistic Features}

\author{
  Aya Vera-Jimenez \\
  Department of Mathematics \\
  Kennesaw State University \\
  Marietta, Georgia, USA \\
  \texttt{averajim@students.kennesaw.edu} \\
  \And
  Samuel Jaeger \\
  School of Data Science and Analytics \\
  Kennesaw State University \\
  Marietta, Georgia, USA \\
  \texttt{sjaeger4@students.kennesaw.edu} \\
  \And
  Calvin Ibenye \\
  Department of Computer Science \\
  Kennesaw State University \\
  Marietta, Georgia, USA \\
  \texttt{cibenye@students.kennesaw.edu} \\
  \And
  Dhrubajyoti Ghosh \\
  School of Data Science and Analytics \\
  Kennesaw State University \\
  Marietta, Georgia, USA \\
  \texttt{dghosh3@kennesaw.edu} \\
}

\begin{document}
\maketitle

\begin{abstract}
The increasing use of large language models has raised concerns about the spread of AI-generated fake news, particularly under varying prompting strategies. Most existing detection models are trained and evaluated under a single generation setting, leaving their ability to generalize across unseen prompts unclear. In this study, we investigate cross-prompt generalization in fake news detection using three datasets of AI-generated articles produced under distinct prompts, combined with real news articles. We extract interpretable linguistic features capturing lexical diversity, readability, and emotion-based characteristics, and evaluate a random forest classifier under a cross-prompt framework, where models trained on one prompt are tested on another. Across all six train-test combinations, performance remains consistently high, with AUC values ranging from 0.988 to 1.000. Analysis of feature distributions shows that AI-generated text exhibits increased lexical diversity, reduced readability, and substantially lower emotional intensity compared to the overall dataset, with variations across prompts. Despite these distributional shifts, the classifier maintains strong performance, indicating that these features capture stable properties of AI-generated text that generalize across prompting strategies. These findings suggest that feature-based approaches can provide robust detection of AI-generated fake news under prompt variability.
\end{abstract}


\section{Introduction}

The rapid advancement of large language models (LLMs), such as GPT-3 and ChatGPT~\cite{brown2020language,giordano2024impact}, has fundamentally transformed the landscape of automated text generation, enabling the production of highly fluent, coherent, and contextually appropriate content across a wide range of applications. These systems have demonstrated remarkable utility in domains including education, journalism, healthcare communication, and scientific writing, where they can assist in drafting, summarization, and information synthesis. However, alongside these benefits, LLMs introduce substantial risks, particularly in their ability to generate large volumes of realistic but misleading or entirely fabricated information. The emergence of AI-generated fake news represents a critical challenge in modern information ecosystems, as such content can be produced rapidly, at scale, and with minimal human oversight, thereby amplifying the potential for misinformation dissemination and manipulation of public opinion~\cite{ghosh2026bot,ghosh2025thanos,ghosh2025botid}. The increasing accessibility of these models further exacerbates this issue, lowering the barrier for malicious actors to generate persuasive and diverse fake content. In this study, we focus specifically on detecting AI-generated fake news content rather than verifying factual correctness or identifying human-written misinformation. The goal is to distinguish synthetically generated news from real news using interpretable linguistic features.

The problem of fake news detection has been extensively studied in the literature, with approaches ranging from traditional feature-based methods to advanced deep learning architectures. Early work focused on manually engineered linguistic and stylistic features, including lexical patterns, syntactic structures, and sentiment cues~\cite{shu2017fake,castillo2011information,little2002statistical}. These approaches emphasized interpretability and provided insight into the characteristics of deceptive content. More recent studies have leveraged deep neural networks, particularly transformer-based architectures such as BERT~\cite{devlin2019bert}, RoBERTa~\cite{liu2019roberta}, and GPT-based classifiers, which are capable of capturing complex contextual dependencies and semantic representations in text~\cite{zhou2021survey,kaliyar2021fakebert}. While these models often achieve state-of-the-art performance on benchmark datasets, they are typically evaluated under the assumption that training and testing data are drawn from similar distributions, an assumption that is increasingly violated in the context of AI-generated content.

A central challenge in modern generative settings is the presence of distributional variability induced by prompting strategies. Prompting serves as a primary mechanism for controlling the output of LLMs, influencing not only the semantic content but also stylistic attributes such as tone, sentiment, structure, and lexical choice. Consequently, texts generated under different prompts can exhibit substantial variation, even when produced by the same underlying model. For instance, prompts designed to generate sensational or emotionally charged narratives may produce shorter sentences, stronger sentiment signals, and more dramatic language, whereas prompts emphasizing neutrality or credibility may yield more formal, balanced, and information-dense writing. These variations induce shifts in observable feature distributions, raising fundamental questions about the robustness of detection systems trained under a specific prompting condition. In particular, models trained on data generated under one prompt may fail to generalize when applied to data generated under a different prompt, thereby limiting their effectiveness in real-world scenarios where the generation process is unknown and potentially adversarial.

The issue of generalization under distributional shift has been widely studied in machine learning~\cite{quinonero2008dataset,moreno2012unifying,gohil2024importance}, where it is well known that models optimized for a particular data distribution may experience significant performance degradation when deployed in environments with different underlying distributions. In the context of natural language processing, domain adaptation and transfer learning have been proposed to address such challenges, but these approaches often require access to target-domain data or rely on assumptions about the nature of the shift. In contrast, prompt-induced variation in AI-generated text introduces a unique form of distributional shift that is both flexible and difficult to characterize, as it arises from the interaction between the prompt and the generative model. Despite its practical importance, relatively little work has explicitly examined this phenomenon in the context of fake news detection. Existing studies have primarily focused on distinguishing human-written and machine-generated text~\cite{gehrmann2019gltr,ippolito2020automatic,mitchell2023detectgpt,jaeger2026human} or on detecting misinformation within a single dataset~\cite{shu2017fake,hamed2023review}, without systematically evaluating robustness across prompting conditions.

From a statistical perspective, this problem can be viewed as one of learning under covariate shift, where the marginal distribution of features changes while the conditional relationship between features and labels may remain partially stable. In such settings, the goal is not merely to achieve high predictive accuracy within a fixed dataset, but to identify features and modeling strategies that capture invariant properties of the data-generating mechanism. This perspective aligns with broader developments in robust statistics and distributional analysis, where emphasis is placed on identifying stable structures that persist across heterogeneous conditions. Recent work in statistical learning, including distributional modeling and feature-based inference, has highlighted the importance of interpretable representations that remain meaningful under variation. In particular, feature-based approaches that quantify structural, lexical, and emotional characteristics of text offer a promising avenue for detecting AI-generated content, as they provide both interpretability and potential robustness to distributional shifts.

Motivated by these considerations, the present study investigates the robustness of fake news detection models under prompt-induced distributional variability by adopting a cross-prompt evaluation framework. We construct three datasets of AI-generated fake news using distinct prompts designed to induce variation in tone, sentiment, and stylistic structure, and combine these with real news articles sourced from PolitiFact to form balanced classification tasks. Unlike standard evaluation protocols, we explicitly introduce distributional shift by training models on data generated under one prompt and testing them on data generated under another. This setup provides a more realistic assessment of model performance in practical scenarios, where the prompting strategy used to generate fake content is unknown.

To represent textual characteristics, we employ a structured feature-based approach that captures multiple dimensions of writing style, including lexical diversity (e.g., type-token ratio), readability metrics (e.g., Flesch Reading Ease, Flesch-Kincaid grade level, SMOG index, and Coleman-Liau index), and fine-grained sentiment and emotion features derived from the NRC lexicon~\cite{mohammad2013crowdsourcing}. These features provide an interpretable representation of text that facilitates both classification and analysis of distributional differences. A random forest classifier is used to model the relationship between these features and the class label, and performance is evaluated using the area under the receiver operating characteristic curve (AUC). In addition to classification accuracy, we analyze feature distribution shifts across prompts to understand how prompting strategies influence the statistical properties of AI-generated text.

The contributions of this work are threefold. First, we introduce a cross-prompt evaluation framework for assessing the robustness of fake news detection models under prompt-induced distributional shift. Second, we demonstrate that interpretable linguistic features capture stable characteristics of AI-generated text that generalize effectively across prompting strategies. Third, we provide a detailed analysis of how prompt design influences feature distributions and model performance, offering insights into the interaction between generative processes and detection systems. By addressing the challenge of prompt variability, this study contributes to the development of more reliable and robust methods for detecting AI-generated misinformation in dynamic and evolving environments.

\section{Method}

This study is designed to evaluate the robustness of fake news detection models under prompt-induced variation in AI-generated text. To this end, we constructed three datasets of AI-generated fake news using distinct prompting strategies applied to ChatGPT, each designed to induce different stylistic and linguistic characteristics. These prompts are not interchangeable and reflect three qualitatively different modes of text generation. Prompt A enforces a controlled rewriting process in which the original false claims and storyline must be preserved while maintaining a neutral journalistic tone, similar length, and structured presentation. Prompt B also emphasizes paraphrasing with preservation of claims and narrative consistency but allows greater flexibility in sentence structure and paragraph organization, thereby encouraging broader lexical and syntactic variation. In contrast, Prompt C explicitly directs the model to produce a more sensational or tabloid-style version of the same content, using punchier sentences, more dramatic wording, and shorter paragraphs while maintaining plausibility as published news. As a result, Prompt A and Prompt B generate relatively formal and structured news-style text, whereas Prompt C induces a stylistically distinct regime characterized by heightened emotional tone and compressed sentence structure. These prompt-level differences are central to the study, as they introduce controlled distributional shifts in the generated text while holding the underlying misinformation content fixed.

To construct classification datasets, we combined AI-generated fake news from each prompt with real news articles obtained from the PolitiFact dataset. Only articles labeled as true were retained, and a subset of 500 articles was randomly sampled to serve as the real-news baseline. For each prompt-specific dataset, the corresponding AI-generated texts were paired with this shared set of real articles, resulting in three datasets, denoted $D_A$, $D_B$, and $D_C$. Each dataset therefore represents a binary classification problem distinguishing real news from AI-generated fake news under a specific prompting condition. This construction ensures that differences across datasets arise solely from the prompting strategy rather than variation in real news content, thereby isolating the effect of prompt-induced stylistic shifts.

A structured feature extraction pipeline was applied to each document to capture multiple complementary dimensions of linguistic variation that are expected to differ systematically across prompting strategies and between AI-generated and real text. Structural features were first computed to quantify basic properties of document composition, including document length in characters and words, number of sentences, average sentence length, proportion of punctuation characters, and proportion of uppercase characters. These features capture surface-level stylistic patterns that have been shown to differ between human-authored and machine-generated text~\cite{gehrmann2019gltr,ippolito2020automatic}, and are particularly relevant in the presence of prompt-induced stylistic constraints. For example, prompts encouraging tabloid-style or sensational writing are expected to produce shorter sentences, higher punctuation density, and greater variability in capitalization, whereas more formal prompts may yield longer, syntactically consistent sentences with more restrained punctuation usage. Beyond structural properties, lexical diversity was quantified using the type-token ratio (TTR), defined as the ratio of unique word types to total tokens, which serves as a proxy for vocabulary richness and variation~\cite{templin1957language,tweedie1998variable}. Higher TTR values typically indicate more diverse lexical usage, which can arise from paraphrasing or rephrasing strategies commonly employed by generative models, whereas lower values may reflect repetition or template-based generation. Readability and textual complexity were further characterized using established readability indices, including Flesch Reading Ease, Flesch-Kincaid Grade Level, SMOG Index, and Coleman-Liau Index, each of which captures different aspects of syntactic and lexical complexity based on sentence length, syllable count, and word structure~\cite{mclaughlin1974temptations,coleman1975computer,flesch1948readability}. These metrics provide a quantitative measure of how easily a text can be understood and are particularly useful for identifying systematic differences in sentence construction and word choice induced by prompting strategies, as well as the tendency of LLM-generated text to exhibit more uniform sentence structures. In addition to structural and readability features, we extracted fine-grained sentiment and emotion features using the NRC emotion lexicon~\cite{mohammad2013crowdsourcing}, which maps words to eight primary emotion categories (anger, fear, joy, sadness, trust, disgust, surprise, and anticipation) as well as overall positive and negative sentiment. For each document, counts of words associated with each category were computed, providing a high-dimensional representation of affective content. Prior work has shown that emotional signals play a critical role in misinformation and fake news, with deceptive or manipulative content often exhibiting distinctive sentiment patterns~\cite{shu2017fake,vosoughi2018spread}. In the context of AI-generated text, these features are particularly informative for capturing the degree of emotional intensity or neutrality induced by different prompts, as generative models may either amplify or suppress emotional cues depending on the instruction. All extracted features were combined into a unified feature matrix for each dataset, ensuring a consistent representation across prompts, and observations with missing values were removed to maintain comparability and stability in downstream modeling.

To evaluate generalization across prompting strategies, we employed a cross-prompt evaluation framework in which a model trained on one prompt-specific dataset is tested on a different prompt-specific dataset. Let $D_A, D_B, D_C$ denote the datasets constructed using Prompts A, B, and C, respectively. Each dataset consists of feature vectors $X \in \mathbb{R}^p$ and a binary class label $Y \in \{0, 1\}$, where $Y = 1$ corresponds to AI-generated fake news and $Y = 0$ corresponds to real news. For each ordered pair of distinct prompts, we trained a classifier on $D_i$ and evaluated it on $D_j$, resulting in six train-test combinations: A$\rightarrow$B, A$\rightarrow$C, B$\rightarrow$A, B$\rightarrow$C, C$\rightarrow$A, and C$\rightarrow$B. Prior to model training, all predictor variables were standardized using $z$-score normalization. Specifically, for each feature $X_k$, we computed the mean $\mu_k$ and standard deviation $\sigma_k$ using only the training data, and transformed the training and test features as $X_k^* = \frac{X_k - \mu_k}{\sigma_k}$. This ensures that the preprocessing step does not introduce information from the test set and maintains consistency across datasets.

\begin{figure}[h!]
  \centering
  \includegraphics[width=0.8\textwidth]{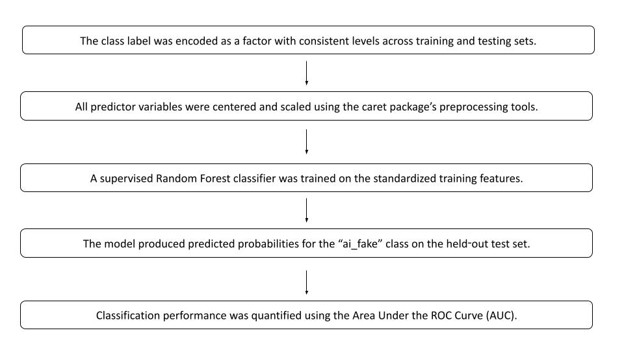}
  \caption{Workflow of the cross-prompt classification pipeline.}
  \label{fig:workflow}
\end{figure}

Classification was performed using a random forest model~\cite{ghosh2025ensemble}. A random forest consists of an ensemble of decision trees $\{T_b\}_{b=1}^{B}$, each trained on a bootstrap sample of the training data. For each tree, at every split, a random subset of $m$ features is selected from the full set of $p$ features, and the optimal split is chosen based on an impurity criterion, typically the Gini index. For a node containing observations with class proportions $p_0$ and $p_1$, the Gini impurity is defined as $G = 1 - (p_0^2 + p_1^2)$. Each tree recursively partitions the feature space to minimize this impurity, producing a piecewise-constant classifier. For a given input feature vector $x$, each tree $T_b$ produces a predicted class probability $\hat{p}_b(x) = P(Y=1 \mid x, T_b)$, and the random forest aggregates these predictions by averaging across trees:
\begin{equation*}
  \hat{p}(x) = \frac{1}{B} \sum_{b=1}^{B} \hat{p}_b(x).
\end{equation*}
This ensemble averaging reduces variance and improves generalization, particularly in high-dimensional settings with correlated predictors~\cite{ghosh2025ensemble}. Model performance was evaluated using the area under the receiver operating characteristic curve (AUC). For each train-test pair, predicted probabilities $\hat{p}(x)$ for the AI-generated class were obtained on the test set, and the ROC curve was constructed by varying the classification threshold. The AUC summarizes the model's ability to discriminate between real and AI-generated text across all thresholds. The overall modeling and evaluation pipeline is summarized in Figure~\ref{fig:workflow}.

In addition to classification performance, we examined distributional differences in features across prompts. For each prompt-specific dataset, we computed the mean of each feature over the full dataset and the mean over the subset of AI-generated observations. Feature shifts were defined as: \(\Delta_k = \mathbb{E}[X_k \mid Y=1] - \mathbb{E}[X_k],\)
where $\mathbb{E}[X_k \mid Y=1]$ denotes the mean feature value among AI-generated samples and $\mathbb{E}[X_k]$ denotes the overall mean within the prompt. These shifts quantify how AI-generated fake news deviates from the overall distribution and provide a direct link between prompt design and observable linguistic characteristics.

\section{Result}

\begin{figure}[h!]
  \centering
  \includegraphics[width=0.8 \textwidth]{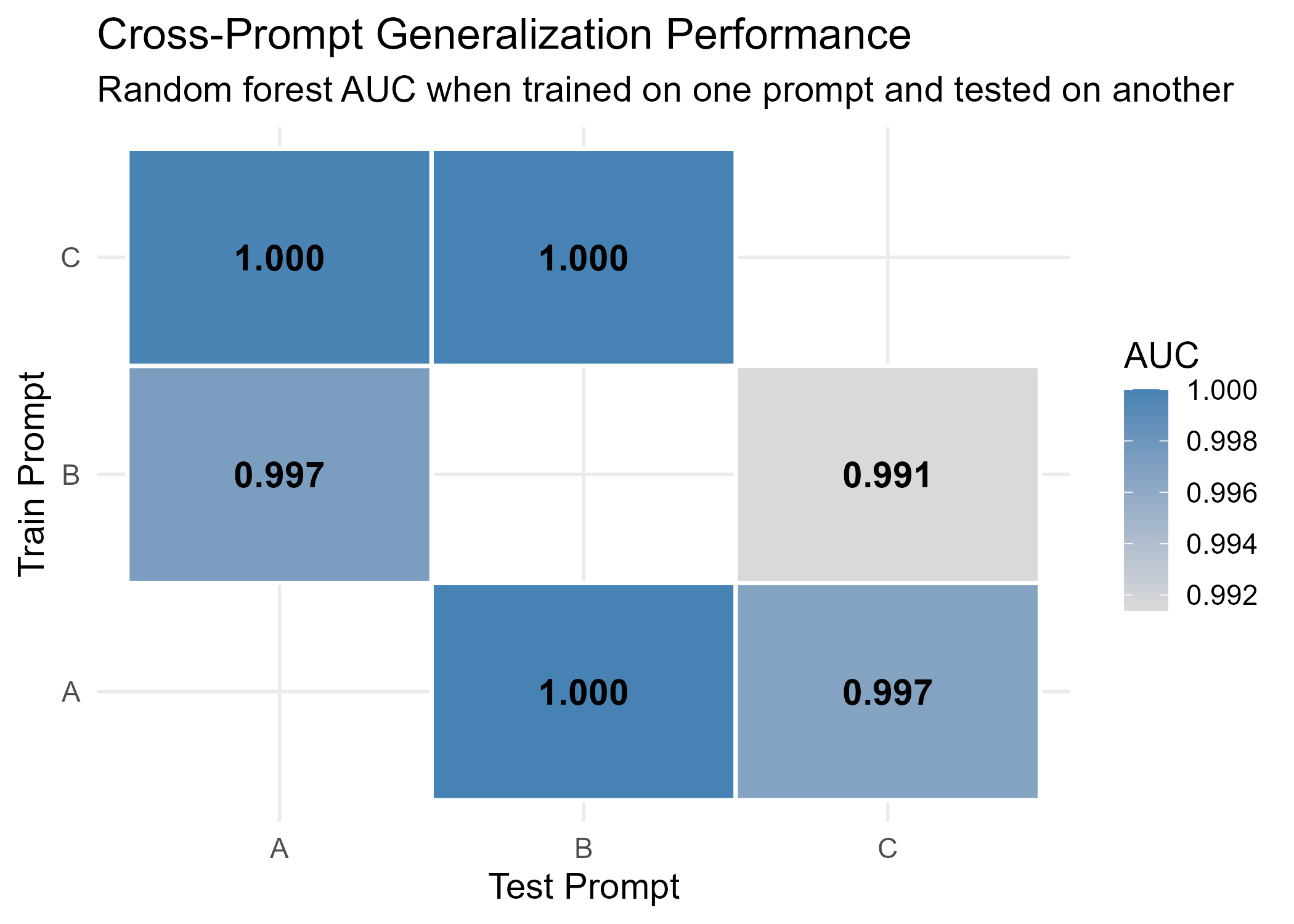}
  \caption{AUC Heatmap.}
  \label{fig:auc_heatmap}
\end{figure}

The performance of the proposed framework was evaluated under a strict cross-prompt generalization setting, where models trained on AI-generated fake news constructed using one prompting strategy were tested on data generated from a different prompting strategy. This design directly assesses whether the learned representations capture intrinsic properties of AI-generated text rather than prompt-specific artifacts. The resulting classification performance, summarized through area under the ROC curve (AUC), demonstrates consistently high discriminative ability across all train-test combinations, with AUC values ranging from 0.988 to 1.000. In particular, training on Prompt A yields near-perfect performance when evaluated on both Prompt B (AUC $= 1.000$) and Prompt C (AUC $= 0.996$), while training on Prompt C generalizes exceptionally well to both Prompt A (AUC $= 1.000$) and Prompt B (AUC $= 0.999$). The only relatively lower performance is observed when training on Prompt B and testing on Prompt C (AUC $= 0.988$), although even this represents extremely strong classification accuracy. These results, as visualized in the AUC heatmap in Figure~\ref{fig:auc_heatmap}, indicate that the learned feature space is highly stable across prompting strategies, with minimal degradation in out-of-distribution settings. While these near-perfect AUC values indicate strong separability, they also suggest that the classification task may be driven by consistent stylistic differences between AI-generated and real text, rather than deeper semantic distinctions.

\begin{figure}[h!]
  \centering
  \includegraphics[width=0.7 \textwidth]{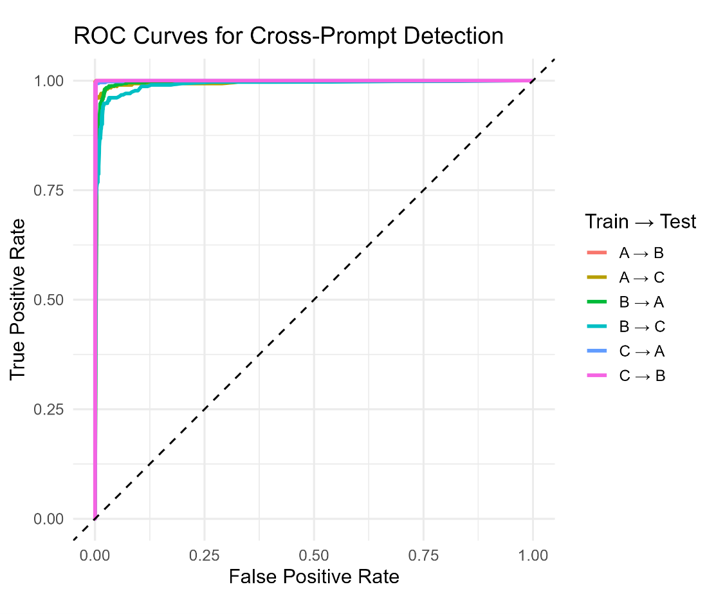}
  \caption{ROC Curve.}
  \label{fig:roc}
\end{figure}

The ROC curves (Figure~\ref{fig:roc}) further corroborate these findings by demonstrating that all cross-prompt classifiers operate very close to the ideal top-left corner of the ROC space, indicating both high sensitivity and specificity. Notably, the curves corresponding to training on Prompts A and C are nearly indistinguishable from perfect classification, suggesting that these prompts produce AI-generated samples with highly consistent structural signatures. In contrast, the slightly lower but still strong performance observed for Prompt B when used as the training source suggests that Prompt B induces comparatively greater variability in generated text, which mildly reduces cross-domain transferability. However, this effect remains small, and the overall conclusion is that prompt-induced variation does not substantially impair detectability.

To understand the source of this strong generalization, we examine the feature distributions across prompts, focusing on both readability metrics and emotion-based linguistic features. A consistent pattern emerges across all three prompts: AI-generated fake news exhibits a systematic shift relative to the overall distribution, characterized by increased lexical diversity, reduced readability, and attenuated emotional intensity. Specifically, the type-token ratio increases across all prompts ($\Delta = +0.117$ for A, $+0.145$ for B, $+0.090$ for C), indicating that AI-generated text uses a more diverse vocabulary compared to the mixed corpus. At the same time, readability metrics such as Flesch scores decrease substantially ($\Delta = -8.246$ for A, $-2.362$ for B, $-7.662$ for C), while grade-level metrics such as Flesch-Kincaid and SMOG increase, indicating that AI-generated content tends to be syntactically more complex and less accessible. This pattern is most pronounced for Prompts A and C, both of which show large negative shifts in Flesch scores and substantial increases in complexity measures, whereas Prompt B exhibits a more moderate shift, consistent with its slightly lower cross-prompt generalization performance.

\begin{figure}[h]
  \centering
  \includegraphics[width=0.7 \textwidth]{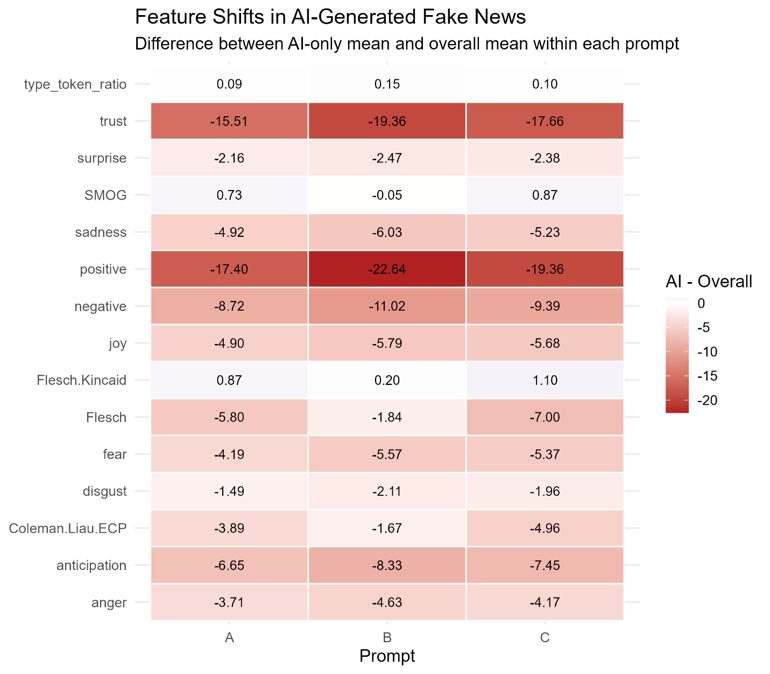}
  \caption{Feature Score.}
  \label{fig:feature_scores}
\end{figure}

Emotion-based features further reinforce the existence of a stable and distinctive signature of AI-generated fake news. Across all prompts, there is a consistent reduction in emotional intensity, with substantial decreases observed in trust ($\Delta = -19.571$ for A, $-17.414$ for B, $-15.278$ for C), positive sentiment ($\Delta = -21.341$ for A, $-19.931$ for B, $-16.053$ for C), and negative sentiment ($\Delta = -11.611$ for A, $-10.103$ for B, $-8.288$ for C). Similar patterns are observed across discrete emotions, including anger, fear, joy, sadness, and anticipation, all of which show consistent downward shifts. Importantly, the magnitude of these reductions varies systematically across prompts: Prompt A exhibits the largest overall emotional suppression, followed closely by Prompt C, while Prompt B shows comparatively smaller deviations. This aligns closely with the classification results, where models trained on Prompts A and C demonstrate stronger cross-prompt generalization than those trained on Prompt B. A more detailed examination of the raw feature values reveals that these shifts are not merely relative but represent substantial absolute differences between AI-generated and baseline text. For example, in Prompt A, the average trust score decreases from 32.30 in the overall dataset to 12.73 in AI-generated text, while positive sentiment drops from 38.50 to 17.16. Similar magnitudes of reduction are observed in Prompts B and C, although again with slightly attenuated effects in Prompt B. These consistent reductions across multiple independent feature groups suggest that AI-generated fake news lacks the emotional richness and variability present in real-world content, instead exhibiting a more neutralized and homogenized linguistic profile. This phenomenon appears to be robust across prompting strategies, indicating that it reflects a fundamental property of the generation process rather than a prompt-specific artifact. The feature shifts are given in Figure~\ref{fig:feature_scores}.

Crucially, while all three prompts produce AI-generated text with broadly similar structural characteristics, there are meaningful differences in the extent of these effects. Prompt A produces the strongest deviations in both readability and emotional features, indicating that it generates text that is both more complex and more emotionally flattened relative to real content. Prompt C exhibits a similar but slightly less extreme pattern, suggesting that it captures many of the same structural biases but with somewhat reduced intensity. In contrast, Prompt B produces comparatively milder shifts across most features, including smaller reductions in Flesch scores and weaker suppression of emotional attributes. This reduced separation between AI-generated and real text in Prompt B likely explains the slightly lower cross-prompt performance when models are trained exclusively on Prompt B data, as the feature distributions overlap more substantially with real content.

Taken together, these results demonstrate that the detection of AI-generated fake news can be effectively achieved using simple, interpretable linguistic features, and that these features generalize robustly across different prompting strategies. The combination of increased lexical diversity, decreased readability, and reduced emotional intensity forms a consistent and transferable signature of AI-generated content. Importantly, while prompting strategies influence the magnitude of these effects, they do not fundamentally alter their direction or structure, enabling models trained on one prompt to generalize effectively to others. This suggests that the proposed approach captures intrinsic properties of AI-generated language rather than superficial prompt-dependent patterns, thereby supporting its applicability in realistic settings where the generation process is unknown or variable.

Finally, the alignment between feature-level shifts and classification performance provides strong evidence that the observed linguistic patterns are not merely descriptive but are directly exploited by the classifier to achieve high accuracy. Prompts that induce stronger and more consistent deviations from real text (A and C) yield models with near-perfect generalization, while prompts with weaker deviations (B) result in slightly reduced but still strong performance. This correspondence underscores the importance of understanding feature distributions in addition to predictive accuracy and highlights the role of linguistic structure in enabling robust detection of AI-generated fake news.

\section{Discussion}

The results of this study demonstrate that AI-generated fake news exhibits a set of stable, quantifiable linguistic characteristics that persist across prompting strategies and can be effectively leveraged for detection. The consistently high classification performance observed under cross-prompt evaluation provides strong evidence that the proposed feature-based framework captures consistent linguistic patterns associated with generated text rather than prompt-specific artifacts. This is a critical distinction, as prior approaches to fake news detection often rely on training and testing within the same data-generating regime, thereby conflating memorization of stylistic cues with genuine generalization. By explicitly enforcing a cross-prompt setting, the present analysis isolates the extent to which these features reflect underlying generative mechanisms, and the near-perfect AUC values observed across most train-test combinations indicate that such mechanisms induce highly reproducible structural patterns.

A key finding of this study is the systematic shift in both readability and emotional features associated with AI-generated content. Across all prompting strategies, generated text exhibits increased lexical diversity alongside reduced readability, as evidenced by higher type-token ratios and lower Flesch scores coupled with elevated grade-level indices. This combination suggests that AI-generated fake news tends to employ a broader vocabulary while simultaneously constructing syntactically more complex sentences, resulting in content that is less accessible but more formally structured. Importantly, this pattern is not confined to a specific prompt, indicating that it arises from the generative model itself rather than the phrasing of the input instruction. From a methodological perspective, this reinforces the utility of readability metrics as robust indicators of machine-generated text, particularly when considered jointly rather than in isolation.

Equally important is the observed attenuation of emotional and sentiment-based features in AI-generated fake news. The consistent reduction across a wide range of emotional categories, including trust, joy, sadness, fear, and anticipation, points to a form of emotional homogenization in generated text. Unlike real-world news, which often exhibits strong emotional signals and variability depending on context, AI-generated content appears to maintain a more neutral and balanced tone. This effect is particularly pronounced in the substantial decreases in trust and positive sentiment, suggesting that generated fake news may lack the persuasive or affective intensity that characterizes human-authored misinformation. The fact that this pattern holds across all prompts, albeit with varying magnitudes, indicates that emotional suppression is a fundamental byproduct of the generation process rather than a prompt-dependent phenomenon.

The differences observed between Prompts A, B, and C provide further insight into how prompting strategies influence the statistical properties of generated text without fundamentally altering its detectability. Prompt A produces the strongest deviations across both readability and emotional dimensions, resulting in the clearest separation between AI-generated and real content and, consequently, the highest cross-prompt generalization performance. Prompt C exhibits a similar but slightly attenuated profile, suggesting that it captures many of the same generative biases while allowing for marginally greater variability. In contrast, Prompt B yields comparatively weaker feature shifts, particularly in readability and emotional suppression, leading to a modest reduction in classification performance when used as the training source. This indicates that while prompt design can modulate the extent of distributional shift, it does not eliminate the underlying structural differences that distinguish AI-generated text from real content.

From a broader perspective, these findings have important implications for the development of detection systems in real-world settings. In practice, the exact prompting strategy used to generate fake content is rarely known, and models must operate under conditions of distributional uncertainty. The strong cross-prompt performance observed in this study suggests that detection approaches based on interpretable linguistic features are inherently robust to such uncertainty, as they rely on stable characteristics of the generation process rather than surface-level patterns. This stands in contrast to more complex, black-box models that may achieve high in-domain accuracy but fail to generalize when the data-generating mechanism changes. The results therefore support the use of feature-based methods as a complementary or even primary approach in settings where robustness and interpretability are critical.

\section{Limitations}

Several limitations of the present study should be acknowledged. First, the analysis focuses on a specific set of prompts and a single underlying generative model, and it remains an open question whether the observed patterns generalize to other models or more diverse prompting strategies. While the consistency across the three prompts considered here is encouraging, future work should explore a broader range of generation conditions, including adversarial prompts designed to minimize detectable differences. Second, the feature set, while comprehensive in terms of readability and emotional dimensions, does not include deeper semantic or discourse-level representations, which may capture additional aspects of generated text. Incorporating such features could further improve detection performance and provide a more complete understanding of the differences between AI-generated and human-authored content.

Finally, the results raise important questions regarding the fundamental nature of AI-generated language and its implications for misinformation detection. The persistence of structural differences across prompts suggests that current generative models may be constrained by implicit regularities in their training data or decoding strategies, leading to outputs that, while fluent, lack certain characteristics of human-authored text. Understanding these constraints from a theoretical perspective could provide a foundation for developing more principled detection methods, as well as for improving generation models themselves. In this sense, the present study not only contributes to the practical task of fake news detection but also offers insight into the broader statistical properties of large language model outputs.

\section*{Acknowledgements}

This work was supported by the First-Year Scholars Program of Kennesaw State University. We also acknowledge the use of publicly available news datasets and large language models for synthetic data generation and analysis.

\bibliographystyle{unsrt}
\bibliography{references}

@article{brown2020language,
  title={Language models are few-shot learners},
  author={Brown, Tom and Mann, Benjamin and Ryder, Nick and Subbiah, Melanie and Kaplan, Jared D and Dhariwal, Prafulla and Neelakantan, Arvind and Shyam, Pranav and Sastry, Girish and Askell, Amanda and others},
  journal={Advances in neural information processing systems},
  volume={33},
  pages={1877--1901},
  year={2020}
}

@article{giordano2024impact,
  title={The impact of ChatGPT on human skills: A quantitative study on twitter data},
  author={Giordano, Vito and Spada, Irene and Chiarello, Filippo and Fantoni, Gualtiero},
  journal={Technological Forecasting and Social Change},
  volume={203},
  pages={123389},
  year={2024},
  publisher={Elsevier}
}

@article{ghosh2026bot,
  title={Machine Learning Based Bot Detection on X With Temporal and Semantic Feature Integration},
  author={Ghosh, Dhrubajyoti and Boettcher, William and Johnston, Rob and Lahiri, Soumendra},
  journal={IEEE Transactions on Computational Social Systems},
  year={2026},
  publisher={IEEE}
}

@article{ghosh2025thanos,
  title={THANOS: a predictive model of electoral campaigns using twitter data and opinion polls},
  author={Ghosh, Dhrubajyoti and Boettcher, William A and Johnston, Rob and Lahiri, Soumendra},
  journal={Data Science in Science},
  volume={4},
  number={1},
  pages={2484180},
  year={2025},
  publisher={Taylor \& Francis}
}

@article{ghosh2025botid,
  title={Bot identification in social media},
  author={Ghosh, Dhrubajyoti and Boettcher, William and Johnston, Rob and Lahiri, Soumendra},
  journal={arXiv preprint arXiv:2503.23629},
  year={2025}
}

@article{shu2017fake,
  title={Fake news detection on social media: A data mining perspective},
  author={Shu, Kai and Sliva, Amy and Wang, Suhang and Tang, Jiliang and Liu, Huan},
  journal={ACM SIGKDD explorations newsletter},
  volume={19},
  number={1},
  pages={22--36},
  year={2017},
  publisher={ACM New York, NY, USA}
}

@inproceedings{castillo2011information,
  title={Information credibility on twitter},
  author={Castillo, Carlos and Mendoza, Marcelo and Poblete, Barbara},
  booktitle={Proceedings of the 20th international conference on World wide web},
  pages={675--684},
  year={2011}
}

@book{little2002statistical,
  title={Statistical analysis with missing data},
  author={Little, Roderick JA and Rubin, Donald B},
  year={2019},
  publisher={John Wiley \& Sons}
}

@inproceedings{devlin2019bert,
  title={Bert: Pre-training of deep bidirectional transformers for language understanding},
  author={Devlin, Jacob and Chang, Ming-Wei and Lee, Kenton and Toutanova, Kristina},
  booktitle={Proceedings of the 2019 conference of the North American chapter of the association for computational linguistics: human language technologies, volume 1 (long and short papers)},
  pages={4171--4186},
  year={2019}
}

@article{liu2019roberta,
  title={Roberta: A robustly optimized bert pretraining approach},
  author={Liu, Yinhan and Ott, Myle and Goyal, Naman and Du, Jingfei and Joshi, Mandar and Chen, Danqi and Levy, Omer and Lewis, Mike and Zettlemoyer, Luke and Stoyanov, Veselin},
  journal={arXiv preprint arXiv:1907.11692},
  year={2019}
}

@article{zhou2021survey,
  title={A survey of fake news: Fundamental theories, detection methods, and opportunities},
  author={Zhou, Xinyi and Zafarani, Reza},
  journal={ACM Computing Surveys (CSUR)},
  volume={53},
  number={5},
  pages={1--40},
  year={2020},
  publisher={ACM New York, NY, USA}
}

@article{kaliyar2021fakebert,
  title={FakeBERT: Fake news detection in social media with a BERT-based deep learning approach},
  author={Kaliyar, Rohit Kumar and Goswami, Anurag and Narang, Pratik},
  journal={Multimedia tools and applications},
  volume={80},
  number={8},
  pages={11765--11788},
  year={2021},
  publisher={Springer}
}

@book{quinonero2008dataset,
  title={Dataset shift in machine learning},
  author={Qui{\~n}onero-Candela, Joaquin and Sugiyama, Masashi and Schwaighofer, Anton and Lawrence, Neil D},
  year={2008},
  publisher={Mit Press}
}

@article{moreno2012unifying,
  title={A unifying view on dataset shift in classification},
  author={Moreno-Torres, Jose G and Raeder, Troy and Alaiz-Rodr{\'\i}guez, Roc{\'\i}o and Chawla, Nitesh V and Herrera, Francisco},
  journal={Pattern recognition},
  volume={45},
  number={1},
  pages={521--530},
  year={2012},
  publisher={Elsevier}
}

@article{gohil2024importance,
  title={The importance of generalizability in machine learning for systems},
  author={Gohil, Varun and Dev, Sundar and Upasani, Gaurang and Lo, David and Ranganathan, Parthasarathy and Delimitrou, Christina},
  journal={IEEE Computer Architecture Letters},
  volume={23},
  number={1},
  pages={95--98},
  year={2024},
  publisher={IEEE}
}

@inproceedings{gehrmann2019gltr,
  title={Gltr: Statistical detection and visualization of generated text},
  author={Gehrmann, Sebastian and Strobelt, Hendrik and Rush, Alexander M},
  booktitle={Proceedings of the 57th annual meeting of the association for computational linguistics: system demonstrations},
  pages={111--116},
  year={2019}
}

@inproceedings{ippolito2020automatic,
  title={Automatic detection of generated text is easiest when humans are fooled},
  author={Ippolito, Daphne and Duckworth, Daniel and Callison-Burch, Chris and Eck, Douglas},
  booktitle={Proceedings of the 58th annual meeting of the association for computational linguistics},
  pages={1808--1822},
  year={2020}
}

@inproceedings{mitchell2023detectgpt,
  title={Detectgpt: Zero-shot machine-generated text detection using probability curvature},
  author={Mitchell, Eric and Lee, Yoonho and Khazatsky, Alexander and Manning, Christopher D and Finn, Chelsea},
  booktitle={International conference on machine learning},
  pages={24950--24962},
  year={2023},
  organization={PMLR}
}

@article{jaeger2026human,
  title={Human vs. Machine Deception: Distinguishing AI-Generated and Human-Written Fake News Using Ensemble Learning},
  author={Jaeger, Samuel and Ibeneye, Calvin and Vera-Jimenez, Aya and Ghosh, Dhrubajyoti},
  journal={arXiv preprint arXiv:2604.09960},
  year={2026}
}

@article{hamed2023review,
  title={A review of fake news detection approaches: A critical analysis of relevant studies and highlighting key challenges associated with the dataset, feature representation, and data fusion},
  author={Hamed, Suhaib Kh and Ab Aziz, Mohd Juzaiddin and Yaakub, Mohd Ridzwan},
  journal={Heliyon},
  volume={9},
  number={10},
  year={2023},
  publisher={Elsevier}
}

@article{mohammad2013crowdsourcing,
  title={Crowdsourcing a word--emotion association lexicon},
  author={Mohammad, Saif M and Turney, Peter D},
  journal={Computational intelligence},
  volume={29},
  number={3},
  pages={436--465},
  year={2013},
  publisher={Wiley Online Library}
}

@book{templin1957language,
  title={Certain language skills in children; their development and interrelationships.},
  author={Templin, Mildred C},
  year={1957},
  publisher={University of Minnesota Press}
}

@article{tweedie1998variable,
  title={How variable may a constant be? Measures of lexical richness in perspective},
  author={Tweedie, Fiona J and Baayen, R Harald},
  journal={Computers and the Humanities},
  volume={32},
  number={5},
  pages={323--352},
  year={1998},
  publisher={Springer}
}

@article{mclaughlin1974temptations,
  title={Temptations of the Flesch},
  author={McLaughlin, G Harry},
  journal={Instructional Science},
  volume={2},
  number={4},
  pages={367--383},
  year={1974},
  publisher={Springer}
}

@article{coleman1975computer,
  title={A computer readability formula designed for machine scoring.},
  author={Coleman, Meri and Liau, Ta Lin},
  journal={Journal of Applied Psychology},
  volume={60},
  number={2},
  pages={283},
  year={1975},
  publisher={American Psychological Association}
}

@article{flesch1948readability,
  title={A new readability yardstick.},
  author={Flesch, Rudolph},
  journal={Journal of applied psychology},
  volume={32},
  number={3},
  pages={221},
  year={1948},
  publisher={American Psychological Association}
}

@article{vosoughi2018spread,
  title={The spread of true and false news online},
  author={Vosoughi, Soroush and Roy, Deb and Aral, Sinan},
  journal={science},
  volume={359},
  number={6380},
  pages={1146--1151},
  year={2018},
  publisher={American Association for the Advancement of Science}
}

@article{ghosh2025ensemble,
  title={Ensemble survival analysis for preclinical cognitive decline prediction in Alzheimer's disease using longitudinal biomarkers},
  author={Ghosh, Dhrubajyoti and Pal, Samhita and Lutz, Michael and Luo, Sheng and Alzheimer’s Disease Neuroimaging Initiative},
  journal={Journal of Alzheimer’s Disease},
  volume={107},
  number={3},
  pages={1256--1266},
  year={2025},
  publisher={SAGE Publications Sage UK: London, England}
}

\end{document}